\pdfoutput=1
\newcommand{\subparagraph}{}

\documentclass[letterpaper, 10 pt, conference]{ieeeconf}  

\IEEEoverridecommandlockouts                              

\title{\LARGE \bf
A Data-Driven Approach for Autonomous Motion Planning and Control in Off-Road Driving Scenarios
}

\usepackage{blindtext, graphicx}
\usepackage{blindtext}
\usepackage{graphicx}
\usepackage[utf8]{inputenc}
\usepackage{graphics} 
\usepackage{epsfig} 
\usepackage{mathptmx} 
\usepackage{amsmath} 
\usepackage{amssymb}  
\usepackage{multicol}
\usepackage{mathtools, cuted}
\usepackage[cmintegrals]{newtxmath}
\usepackage{epstopdf}

\usepackage{helvet}
\usepackage{courier}
\usepackage[tight,footnotesize]{subfigure}
\usepackage{type1cm}
\usepackage{titlesec}
\usepackage{epsfig} 
\usepackage{mathptmx} 
\usepackage{makeidx}         
\usepackage{fmtcount} 
\usepackage{multirow}
\usepackage{algorithmic}
\DeclareMathOperator*{\argmin}{\arg\!\min}

\usepackage{multicol}        
\usepackage[bottom]{footmisc}
\usepackage{caption}
\usepackage{titlesec}
\usepackage{color}

\usepackage{graphics} 
\usepackage{graphicx}
\usepackage{epsfig} 
\usepackage{graphicx}
\usepackage{adjustbox,lipsum}
\usepackage{fancyhdr}

\fancypagestyle{plain}{
  \fancyhf{}
  \fancyhead[C]{Conference on \LaTeX}     
  \fancyfoot[L]{This is a notice}

}

\begin{document}
%
\author{Hossein~Rastgoftar, Bingxin Zhang, ~and~Ella~M.~Atkins
}
\maketitle

\thispagestyle{fancy}
\cfoot{This paper will appear in the 2018 American Control Conference.}

\begin{abstract}
This paper presents a novel data-driven approach for vehicle motion planning and control in off-road driving scenarios. For autonomous off-road driving, environmental conditions impact terrain traversability as a function of weather, surface composition, and slope.  Geographical information system (GIS) and National Centers for Environmental Information datasets are processed to provide this information for interactive planning and control system elements. A top-level global route planner (GRP) defines optimal waypoints using dynamic programming (DP). A local path planner (LPP) computes a desired trajectory between waypoints such that infeasible control states and collisions with obstacles are avoided. The LPP also updates the GRP with real-time sensing and control data. A low-level feedback controller applies feedback linearization to asymptotically track the specified LPP trajectory. Autonomous driving simulation results are presented for traversal of terrains in Oregon and Indiana case studies.

\end{abstract}



%
\IEEEpeerreviewmaketitle

\section{Introduction}
Success in autonomous off-road driving will be assured in part through the use of diverse static and real-time data sources in planning and control decisions.  A vehicle traversing complex terrain over a long distance must define a route that balances efficiency with safety. Off-road driving requires including both global and local metrics in path planning. Algorithms such as A* and D* \cite{duchovn2014path} for waypoint definition and sequencing can be combined with local path planning strategies \cite{latombe2012robot} to guarantee obstacle avoidance given system motion constraints and terrain properties such as slope and surface composition. 

Off-road navigation studies to-date have primarily focused on avoiding obstacles and improving local driving paths. A* \cite{gaw1986minimum} and dynamic programming \cite{zhao2014optimizing} support globally-optimal planning over a discrete grid or pre-defined waypoint set. In Ref.  \cite{karumanchi2010non}, the surface slope is considered during A* search over a grid-based mobility map.  Slope is used to assign feasible traversal velocity constraints that maintain acceptable risk of loss-of-control (e.g., spin-out) or roll-over. In Ref. \cite{chu2012local}, local driving path optimization is studied, while Ref. \cite{chu2015real} applies a Pythagorean Hodograph (PH) \cite{farouki1990pythagorean} cubic curve  to provide a smooth path that avoids obstacles by generating a kinematic graph data structure. Vehicle models that consider continuously-variable (rough) terrain are introduced in Refs. \cite{howard2007optimal}, \cite{shiller1990optimal}, \cite{shiller1991dynamic}, \cite{amar1993modeling} and \cite{bonnafous2001motion}. 


This paper proposes a data-driven approach to autonomous vehicle motion planning and control for off-road driving scenarios. 
The decision-making architecture consists of three layers: (i) Global Route Planning (GRP), (ii) Local Path Planning (LPP), and (iii) Feedback Control (FC). The GRP planning layer assigns optimal waypoints using dynamic programming (DP) \cite{bertsekas1995dynamic, zolfpour2014modeling}. The GRP must rely on cloud and stored database information to define and sequence waypoints beyond onboard sensor range (line-of-sight).  In this paper, DP cost is defined based on realistic weather and geographic data provided by the National Center  for  Environmental  Information  in the National Oceanographic and Atmospheric Administration (NOAA) along with geographical information system (GIS) data. The LPP computes a continuous-time trajectory between optimal waypoints assigned by the GRP. The FC applies a nonlinear controller to asymptotically track desired vehicle trajectory.  To our knowledge, this is the first publication in which NOAA weather and GIS data provide input into autonomous off-road driving decision-making. 

Autonomous off-road driving has been proposed for multiple applications. The DARPA Grand Challenge series and PerceptOR program have led to numerous advances in perception and autonomous driving decision-making.  For example, Ref. \cite{kelly2006toward} proposes a three-tier deliberative, perception, reaction architecture to navigate an off-road cluttered environment with limited GPS availability and changing lighting conditions. A review of navigation or perception systems relevant to agricultural applications is provided in Ref. \cite{mousazadeh2013technical}.  Because agriculture equipment normally operates in open fields with minimal slope, precision driving and maneuvering tends to be more important than evaluating field traversability. Ref. \cite{reina2016lidar} describes how LIDAR and stereo video data can be fused to support off-road vehicle navigation, providing critical real-time traversability information for the area within range of sensors. To augment LIDAR and vision with information on soil conditions, Ref. \cite{gonzalez2017thermal} proposes use of a thermal camera to provide real-time measurements of soil moisture content, which in turn can be used to assess local traversability.  Our paper provides complementary work that incorporates GIS and cloud-based (NOAA) data sources to enable an off-road vehicle planner to build a traversable and efficient route through complex off-road terrain.  Onboard sensors would then provide essential feedback to confirm the planned route is safe and update database-indicated traversability conditions as needed.

A second contribution of this paper is a feedback linearization controller for trajectory tracking over nonlinear surfaces. Most literature on vehicle control and trajectory tracking usually assumes that the car moves on a flat surface \cite{hwan2013optimal, kong2015kinematic, ryu2004integrating, brown2017safe}. In Ref. \cite{brown2017safe}, model predictive controller (MPC) is deployed for trajectory tracking and motion control on flat surfaces.  Ref. \cite{howard2007optimal} introduces a vehicle body frame for motion over a nonlinear surface and realizes velocity with respect to the local coordinate frame. Bases of the body and ground (world) coordinate system are related through Euler angles $\phi$, $\theta$, $\psi$. A first-order kinematic model for motion over a nonlinear surface is presented in Ref. \cite{howard2007optimal}. 

In this paper, we extend the kinematic car model given in Refs. \cite{paden2016survey, kwatny2000nonlinear} by including both position and velocity as control states. The car is modeled by two wheels connected by a rigid bar. Side-slip of the rear car wheel (axle) is presumed zero. Car tangent acceleration (drive torque) and steering rate control inputs are chosen such that the desired trajectory on a nonlinear motion surface is asymptotically tracked.

This paper is organized as follows. Section \ref{Preliminaries} presents background on dynamic programming, deterministic state machines, and ground vehicle (car) kinematics. Section \ref{Methodology} describes the paper's methodology, followed by off-road driving simulation results in Section \ref{Simulation Results}. Section \ref{Conclusion} concludes the paper.

\section{Preliminaries}
\label{Preliminaries}
Three motion planning and control layers are integrated to enable off-road autonomous driving. The top layer is "global route planning" (GRP) using dynamic programming (DP) as reviewed in Section \ref{Dynamic Programming}. The second layer, local path planning (LPP), assigning a continuous-time vehicle trajectory to waypoints sequenced by the GRP via path and speed planning computations. 
The paper defines a deterministic finite state machine (DFSM) \cite{hopcroft2006automata} to specify desired speed along the desired path. Elements of a DFSM are defined in Section \ref{Deterministic Finite State Machine}.
The paper presents a nonlinear feedback control approach for trajectory tracking over an arbitrary motion surface as the inner-loop (third) layer. Motion of the car is expressed in coordinate frames defined in Section \ref{Local and Ground Coordinate Systems} using driving dynamics given in Section \ref{Ego Car Dynamics}.
\subsection{Dynamic Programming}
\label{Dynamic Programming}
A dynamic programming (DP) problem \cite{bertsekas1995dynamic} can be defined by the tuple 
\[
    \left(\mathcal{S}, \mathcal{A}, \mathcal T, \mathcal C\right)
\]
where $\mathcal{S}$ is a set of discrete states with cardinality $N$, and $\mathcal{A}$
is the set of discrete actions with cardinality $n_a$.  Furthermore, $\mathcal{C}:\mathcal{S}\times \mathcal{A}\rightarrow \mathbb{R}$ is the cost function and  $\mathcal{T}:\mathcal{S}\times \mathcal{A}\rightarrow \mathcal{S}$ is a deterministic transition function.



The Bellman Equation defines optimality with respect to action $a^*(s)\in \mathcal{A}$ selected for each state $s$:
\begin{equation}
    \mathcal{V}(s)=\min\limits_{\forall a\in \mathcal{A},\forall s'\in \mathcal{S}}\{C(s,a)+\mathcal{T}(s,a,s')V(s')\}
\end{equation}
where $\mathcal{V}:\mathcal{S}\rightarrow \mathbb{R}$ is the utility or value function. Therefore, 
\begin{equation}
    a^*(s)=\argmin\limits_{\forall a\in \mathcal{A},\forall s'\in \mathcal{S}}\{C(s,a)+\mathcal{T}(s,a,s')V(s')\}
\end{equation}
assigns the optimal action for each state $s\in \mathcal{S}$. Case study simulations in this paper use a traditional value iteration algorithm \cite{puterman2014markov} to solve the Bellman equation.

\subsection{Deterministic Finite State Machine}
\label{Deterministic Finite State Machine}
The behavior of a discrete system can be represented by a directed graph formulated as a \textit{finite state machine} (FSM). In a FSM, nodes represent discrete states of the system and edges assign transitions between states. A FSM is \textit{deterministic} if a unique input signal always returns the same result.  A deterministic finite state machine (DFSM) \cite{hopcroft2006automata} is mathematically defined by the tuple 
\[
\left(\Sigma,\mathcal P, \mathcal F, \Delta, p_0\right)
\]
where $\Sigma $ is a finite set of inputs, $\mathcal P$ is the finite set of states, $\mathcal F$ is the set of terminal states, $\Delta:\mathcal P\times \Sigma \rightarrow \mathcal P$ defines transitions over the DFSM states, and $p_0$ is the initial state.
\begin{figure}
\center
\includegraphics[width=3 in]{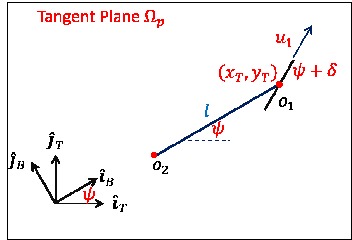}
\caption{Off-road driving coordinate frames.}
\label{carbodyframe}
\end{figure}

\subsection{Motion Kinematics}
\label{Local and Ground Coordinate Systems}
\subsubsection{Ground Coordinate System}
Bases of a ground-fixed or inertial coordinate system are denoted $\hat{\mathbf{i}}_G$, $\hat{\mathbf{j}}_G$, and $\hat{\mathbf{k}}_G$, where $\hat{\mathbf{i}}_G$ and $\hat{\mathbf{j}}_G$ locally point East and North, respectively. Position of the vehicle can be expressed with respect to ground coordinate system $G$ by
\begin{equation}
    \mathbf{r}=x\hat{\mathbf{i}}_G+y\hat{\mathbf{j}}_G+z\hat{\mathbf{k}}_G
\end{equation}
Because the ground coordinate system is stationary, $\dot{\mathbf{i}}_G=\mathbf{0}$, $\dot{\mathbf{j}}_G=\mathbf{0}$, and $\dot{\mathbf{k}}_G=\mathbf{0}$. Velocity and  accelration are assigned by
\begin{equation}
\label{VELACC}
\begin{split}
    \dot{\mathbf{r}}=&\dot{x}\hat{\mathbf{i}}_G+\dot{y}\hat{\mathbf{j}}_G+\dot{z}\hat{\mathbf{k}}_G\\
    \ddot{\mathbf{r}}=&\ddot{x}\hat{\mathbf{i}}_G+\ddot{y}\hat{\mathbf{j}}_G+\ddot{z}\hat{\mathbf{k}}_G\\
\end{split}
\end{equation}

It is assumed that the vehicle (car) moves on a surface
\begin{equation}
\label{zfxy}
\Phi_p=z-f(x,y)=0.
\end{equation}
\subsubsection{Local Coordinate System}
Bases of a local \textit{terrain} coordinate system $T$ are denoted $\hat{\mathbf{i}}_T$, $\hat{\mathbf{j}}_T$, and $\hat{\mathbf{k}}_T$, where $\hat{\mathbf{k}}_T$ is normal to surface $\Phi_p$. Therefore, 
\begin{equation}
\label{kbxy}
\begin{split}
   \hat{\mathbf{k}}_T(x,y)=k_{T,x}(x,y)\hat{\mathbf{i}}_G+k_{T,y}(x,y)\hat{\mathbf{j}}_G+k_{T,z}(x,y)\hat{\mathbf{k}}_G=\dfrac{\bigtriangledown \Phi_p}{\|\bigtriangledown \Phi_p\|}
\end{split}
.
\end{equation}
Bases of the local coordinate system are related to the bases of the ground frame by
\begin{equation}
\label{ibjbkb2ijk}
\begin{split}
    \begin{bmatrix}
    \hat{\mathbf{i}}_T\\
    \hat{\mathbf{j}}_T\\
    \hat{\mathbf{k}}_T\\
    \end{bmatrix}
    =
    \mathcal{R}_{\phi-\theta}
    \begin{bmatrix}
    \hat{\mathbf{i}}_G\\
    \hat{\mathbf{j}}_G\\
    \hat{\mathbf{k}}_G\\
    \end{bmatrix}
\end{split}
,
\end{equation}
where
\[
\begin{split}
\mathcal{R}_{\phi-\theta}=&\mathcal{R}_\phi\mathcal{R}_\theta=\begin{bmatrix}
    \cos\theta&0&-\sin\theta\\
    \sin\phi \sin\theta&\cos\phi&\sin\phi \cos\theta\\
    \cos\phi \sin \theta&-\sin\phi&\cos\phi \cos\theta\\
    \end{bmatrix}
\\
\mathcal{R}_{\phi}=&
\begin{bmatrix}
1&0&0\\
0&\cos\phi&\sin\theta\\
0&-\sin\phi&\cos\phi
\end{bmatrix}
\\
\mathcal{R}_{\theta}=&
\begin{bmatrix}
\cos\theta&0&-\sin\theta\\
0&1&0\\
\sin\theta&0&\cos\theta
\end{bmatrix}
\end{split}
.
\]

Equating the right hand sides of Eq. \eqref{kbxy} and the third row of \eqref{ibjbkb2ijk}, the roll angle $\phi$ and the pitch angle $\theta$ are obtained as follows:
\begin{equation}
\label{phitheta}
\begin{split}
   \phi=&\sin^{-1}\left(-k_{T,y}\right) \\
   \theta=&\tan^{-1}\left(\dfrac{k_{T,x}}{k_{T,z}}\right) \\
\end{split}
.
\end{equation}
$\dot{\phi}$ and $\dot{\theta}$ can be related to $\dot{x}$ and $\dot{y}$ by taking the derivative of Eq. \eqref{phitheta}:
\begin{equation}
\label{DPHDTH}
    \begin{bmatrix}
    \dot{\phi}\\
    \dot{\theta}
    \end{bmatrix}
    =\begin{bmatrix}
    \dfrac{1}{C\phi}&0\\
    0&{C\theta}^2
    \end{bmatrix}
   \begin{bmatrix}
\dfrac{-\partial k_{T,y}}{\partial x}&\dfrac{-\partial k_{T,y}}{\partial y}\\
\dfrac{\dfrac{\partial k_{T,x}}{\partial x}-\dfrac{\partial k_{T,z}}{\partial x}}{k_{T,z}^2}&\dfrac{\dfrac{\partial k_{T,x}}{\partial y}-\dfrac{\partial k_{T,z}}{\partial y}}{k_{T,z}^2}
\end{bmatrix}
    \begin{bmatrix}
    \dot{x}\\
    \dot{y}
    \end{bmatrix}
    .
\end{equation}
Note that $C\phi$ and $C\theta$ abbreviate $\cos\phi$ and $\cos\theta$, respectively.
\subsubsection{Body Coordinate System}
The bases of the vehicle body coordinate system $B$, denoted $\hat{\mathbf{i}}_B$, $\hat{\mathbf{j}}_B$, and $\hat{\mathbf{k}}_B$, can be related to $\hat{\mathbf{i}}_T$, $\hat{\mathbf{j}}_T$ and $\hat{\mathbf{k}}_T$ by
\begin{equation}
    \begin{bmatrix}
    \hat{\mathbf{i}}_B\\
    \hat{\mathbf{j}}_B\\
    \hat{\mathbf{k}}_B\\
    \end{bmatrix}
    =
    \mathcal{R}_{\psi}
    \begin{bmatrix}
    \hat{\mathbf{i}}_T\\
    \hat{\mathbf{j}}_T\\
    \hat{\mathbf{k}}_T\\
    \end{bmatrix}
=
    \begin{bmatrix}
    \cos\psi&\sin\psi&0\\
    -\sin\psi&\cos\psi&0\\
    0&0&1
    \end{bmatrix}
    \begin{bmatrix}
    \hat{\mathbf{i}}_T\\
    \hat{\mathbf{j}}_T\\
    \hat{\mathbf{k}}_T\\
    \end{bmatrix},
\end{equation}
where $\psi$ is the yaw or approximate heading angle.
Car angular velocity can be expressed by
\begin{equation}
    \vec{\omega}_B=\vec{\omega}_T+\dot{\psi}\hat{\mathbf{k}}_T,
\end{equation}
where 
\begin{equation}
\label{AngVel}
\begin{split}
    \vec{\omega}_T=&p_T\hat{\mathbf{i}}_T+q_T\hat{\mathbf{j}}_T+r_T\hat{\mathbf{k}}_T\\=&\dot{\phi}\hat{\mathbf{i}}_T+\dot{\theta}\cos\phi\hat{\mathbf{j}}_T-\dot{\theta}\sin\phi\hat{\mathbf{k}}_T
\end{split}
.
\end{equation}

\subsection{{Vehicle Dynamics}}
\label{Ego Car Dynamics}
Fig. \ref{carbodyframe} shows a schematic of vehicle/car configuration in motion plane $\Omega_p$; we assume $\Omega_p$ is tangent to the terrain surface.
The car is modeled by two wheels connected by a rigid bar with length $l$. In the figure, $o_1$ and $o_2$ are the centers of the front and rear tires, respectively. Given 
steering angle $\delta$ 
and car speed $v_T$,  motion dynamics can be expressed by \cite{paden2016survey, kwatny2000nonlinear}
\begin{equation}
\label{EGOCARDYN}
\begin{split}
\dot{\mathbf{r}}=&v_T\left(\cos\delta\hat{\mathbf{i}}_B+\sin\delta\hat{\mathbf{j}}_B\right)\\
\end{split}
.
\end{equation}

\textbf{Motion Constraint}: This work assumes side slip of the rear tire is zero. This assumption can be mathematically expressed by
\begin{equation}
    \mathbf{v}_{{\mathrm{rel}}_{o_2}}\cdot\hat{\mathbf{j}}_B=\mathbf{0},
\end{equation}
where
\begin{equation}
\mathbf{v}_{{\mathrm{rel}}_{o_2}}=\dot{\mathbf{r}}-\vec{\omega}_B\times (-l\hat{\mathbf{i}}_B)
\end{equation}
is the rear tire relative velocity with respect to the local body frame. Therefore 
\begin{equation}
    \dot{\psi}=-\dfrac{1}{l}\left(\dot{{\mathbf{r}}}-l\vec{\omega}_T\times\hat{\mathbf{i}}_B\right)\cdot\hat{\mathbf{j}}_B.
\end{equation}

By taking the time derivative of Eq. \eqref{EGOCARDYN}, acceleration of the car is computed as
\begin{equation}
\begin{split}
\ddot{\mathbf{r}}=&a_T\left(\cos\delta\hat{\mathbf{i}}_B+\sin\delta\hat{\mathbf{j}}_B\right)+v_T\gamma\left(-\sin\delta\hat{\mathbf{i}}_B+\cos\delta\hat{\mathbf{j}}_B\right)\\
+&\vec{\omega}_B\times v_T\left(\cos\delta\hat{\mathbf{i}}_B+\sin\delta\hat{\mathbf{j}}_B\right)    
\end{split}
,
\end{equation}
where $a_T=\dot{v}_T$ and $\gamma=\dot{\delta}$ are the car tangential acceleration and steering rate, respectively. $a_T$ and $\gamma$ are determined by
\begin{equation}
\label{atgamma}
\begin{split}
&  \begin{bmatrix}
    a_T&
    \gamma
    \end{bmatrix}
    ^T
    =
    \\
    &
    \begin{bmatrix}
    \cos\delta&\sin\delta\\
    \dfrac{-\sin\delta}{v_T}&\dfrac{\cos\delta}{v_T}
    \end{bmatrix}
    \begin{bmatrix}
    \hat{\mathbf{i}}_B\cdot\big[ \ddot{\mathbf{r}}-\vec{\omega}_B\times v_T\left(\cos\delta\hat{\mathbf{i}}_B+\sin\delta\hat{\mathbf{j}}_B\right)\big]\\
    \hat{\mathbf{j}}_B\cdot\big[ \ddot{\mathbf{r}}-\vec{\omega}_B\times v_T\left(\cos\delta\hat{\mathbf{i}}_B+\sin\delta\hat{\mathbf{j}}_B\right)\big]\\
    \end{bmatrix}
    .
\end{split}
\end{equation}

The magnitude of vehicle normal force,
\begin{equation}
\label{FN}
    {F}_N=\hat{\mathbf{k}}_B\cdot\left(mg\mathbf{k}_G+m\ddot{\mathbf{r}}\right)
\end{equation}
must be always positive, e.g. $F_N>0,\forall t$. This guarantees that the car never leaves the motion surface. 

\textbf{Remark}: In Eqs. \eqref{EGOCARDYN}, \eqref{atgamma} and \eqref{FN}, $\dot{\mathbf{r}}$ and $\ddot{\mathbf{r}}$ are the car velocity and acceleration expressed with respect to the ground coordinate system (see Eq. \eqref{VELACC}).

\section{Methodology}
\label{Methodology}
This section presents the proposed three-layer planning strategy comprised of Global Route Planning (GRP), Local path planning (LPP), and feedback control (FC). GRP applies DP to assign optimal driving waypoints given terrain navigability (traversability), weather conditions, and driving motion constraints.
 LPP is responsible for computing a trajectory  between consecutive waypoints assigned by DP. Local obstacle information obtained during LPP is applied by FC to track the desired trajectory.
\subsection{Global Route Planning}
\label{Global Route Planning}
\textbf{DP state set $\mathcal S$}: A uniform grid is overlaid onto a local terrain of the United States. Grid nodes are defined by the set $\mathcal{V}$; the node $i\in \mathcal{V}$ is considered as an obstacle if:
\begin{enumerate}
    \item{There exists water at node location $i\in \mathcal{V}$, }
    \item{Node $i\in \mathcal{V}$ contains foliage (trees) or buildings, or}
    \item{There is a considerable elevation difference (steep slope) at node location $i\in \mathcal{V}$.}
\end{enumerate}
Let the set $\mathcal{V}_o$ define obstacle index numbers. Then,
\begin{equation}
    \mathcal S=\mathcal{V}\setminus \mathcal{V}_o
\end{equation}
defines the DP states. We assume that $\mathcal S$ has cardinality $N$, e.g. $\mathcal{S}=\{1,\cdots,N\}$. 

Transition from node $s\in \mathcal S$ to node $s'\in \mathcal S$ is defined by a directed graph.
In-neighbor nodes of node $s\in \mathcal{S}$ are defined by the set 
\begin{equation}
    \mathcal N_s=\{s_1,\cdots,s_{n_s}\},
\end{equation}
where $n_s\leq 8$ is the cardinality of the set $\mathcal N_s$. 

\begin{figure}
\center
\includegraphics[width=3.3 in]{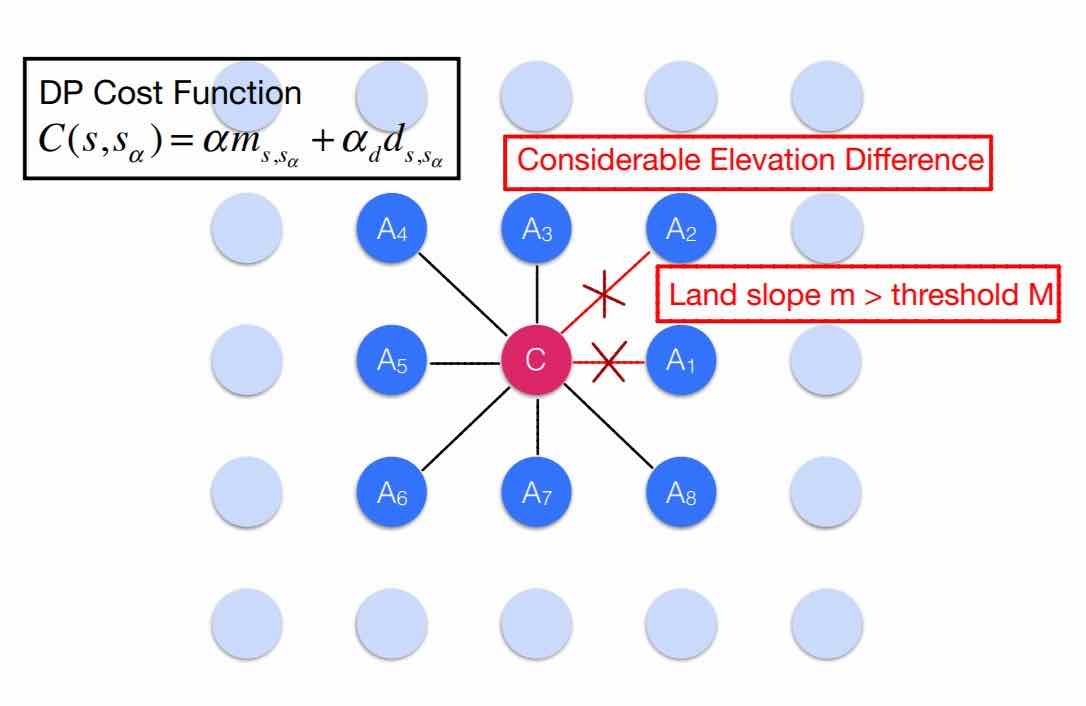}
\caption{Adjacent nodes with minimum cost-to-go are preferred. If there is non-zero elevation difference between two adjacent points choose only actions respecting weather-dependent constraints $M_{d,max}$ or $M_w,max$.}
\label{DPdiagram}
\end{figure}

\textbf{DP Actions}: DP actions are defined by the set 
\begin{equation}
\mathcal{A}=\{1,\cdots, 9\},
\end{equation}
where actions $1$ through $8$ command the vehicle to drive to an adjacent node directly East, Northeast, North, Northwest, West, Southwest, South, and Southeast, respectively. Action $9\in \mathcal{A}$ is the "Stay" command. As shown in Fig. \ref{DPdiagram}, all directions defined by the set $\mathcal{S}$ may not necessarily be reached from every node $s\in \mathcal{S}$. Therefore, actions available at node $s\in \mathcal{S}$ are defined by $\mathcal{A}_s\subset \mathcal{A}$. 

\textbf{Transition Function}: Let $(x_s,y_s)$ and $(x_{s'},y_{s'})$ denote planar positions of nodes $s\in \mathcal{S}$ and $s'\in \mathcal{N}_s$; land slope $m_{s,s'}$ over the straight path connecting $s$ and $s'$ is considered as the criterion for land navigability in this paper. We define $M_{d,max}$ and $M_{w,max}$ as upper bounds for land slope $m_{s,s'}$ for dry and hazardous (wet) surface conditions, respectively, leading to the following constraints:
\begin{itemize}
    \item{In a dry weather condition, $s'\in \mathcal{N}_s$ can be reached from $s\in \mathcal{S}$ only when $m_{s,s'}\leq M_{d,max}$.}
        \item{In a wet weather condition, $s'\in \mathcal{N}_s$ can be reached from $s\in \mathcal{S}$ only when $m_{s,s'}\leq M_{w,max}$.}
\end{itemize}
Suppose that $s_a\in \mathcal{S}$ is the expected outcome state when executing action $a\in \mathcal{A}$ in $s\in \mathcal{S}$. Then, transition function $\mathcal{T}\left(s,a,s'\right)$ is defined as follows:
\begin{equation}
\begin{split}
     &\mathcal{T}\left(s,a,s'\right)=\\
     &
    \begin{cases}
    1&\left(s'=s_a\right)\wedge\left(m_{s,s_a}\leq M_{d,max}\vee m_{s,s_a}\leq M_{w,max}\right)\\
    0&\mathrm{else}.
    \end{cases}
\end{split}
\end{equation}

\textbf{DP cost}: The DP cost at node $s\in \mathcal{S}$ under DP action $a\in \mathcal{A}_s$ is defined by 
\begin{equation}
    C(s,s_a)=\alpha_m\bar{m}_{s,s_a}+\alpha_dd_{s,s_a},
\label{DPequation}
\end{equation}
where $\bar{m}_{s,s_a}$ is the average slope (elevation difference) along the path segment connecting $s\in \mathcal{S}$ and $s_a\in \mathcal{N}_s$.  Also, $d_{s,s_a}$ is the distance between nodes $s\in \mathcal{S}$ and $s_a\in \mathcal{N}_s$.
Note that scaling factors $\alpha_m$ and  $\alpha_d$ are assigned by 
\begin{equation}
\begin{bmatrix}
 \bar{m}&\bar{d}\\
 1&1
\end{bmatrix}
\begin{bmatrix}
\alpha_m\\
\alpha_d
\end{bmatrix}
=
\begin{bmatrix}
1\\
1
\end{bmatrix}
,
\end{equation}
where 
\begin{equation}
\begin{split}
    \bar{m}=&\mathrm{\mathbf{Average}}\big\{m(s,a)\big|s\in \mathcal{S},a\in \mathcal{A}_s\big\}\\
    \bar{d}=&\mathrm{\mathbf{Average}}\big\{d(s,a)\big|s\in \mathcal{S},a\in \mathcal{A}_s\big\}\\
\end{split}
.
\end{equation}

\subsection{Local Path Planning}
The main responsibility of the local path planner (LPP) is to define a desired trajectory between consecutive waypoints assigned by the DP-based GRP. The LPP also might interact with the GRP to share information about dynamically-changing obstacle and environment properties (in future work).  LPP trajectory computation is discussed below.  
\begin{figure}
\center
\includegraphics[width=2.5 in]{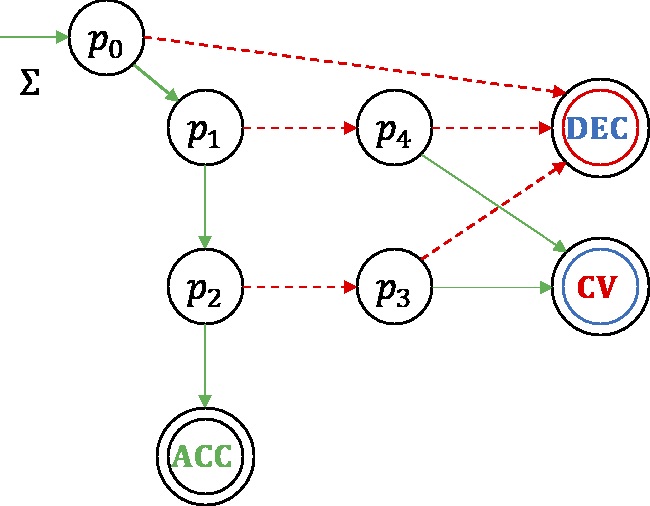}
\caption{Trajectory planning state machine.}
\label{TPSM}
\end{figure}
\subsubsection{Trajectory Planning}
Suppose $(x_{k-1},y_{k-1})$, $(x_{k},y_{k})$, and $(x_{k+1},y_{k+1})$ are $x$ and $y$ components of three consecutive way points, where the path segments connecting these three waypoints are navigable, e.g. the path connecting these three waypoints are obstacle-free. Let 
\begin{equation}
\begin{split}
  \mu_{k-1,k}=&\dfrac{y_{k}-y_{k-1}}{x_{k}-x_{k-1}}\\
  \mu_{k,k+1}=&\dfrac{y_{k+1}-y_k}{x_{k+1}-x_k}\\
\end{split}
,
\end{equation}
then the path segments connecting $(x_{k-1},y_{k-1})$, $(x_{k},y_{k})$, and $(x_{k+1},y_{k+1})$ intersect if $\mu_{k-1,k}\neq \mu_{k,k+1}$. 

\begin{figure}
\centering
\subfigure[$\mu_{k-1,k}= \mu_{k,k+1}$]{\includegraphics[width=0.48\linewidth]{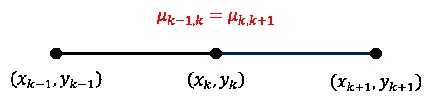}}
\subfigure[$\mu_{k-1,k}\neq \mu_{k,k+1}$]{\includegraphics[width=0.48\linewidth]{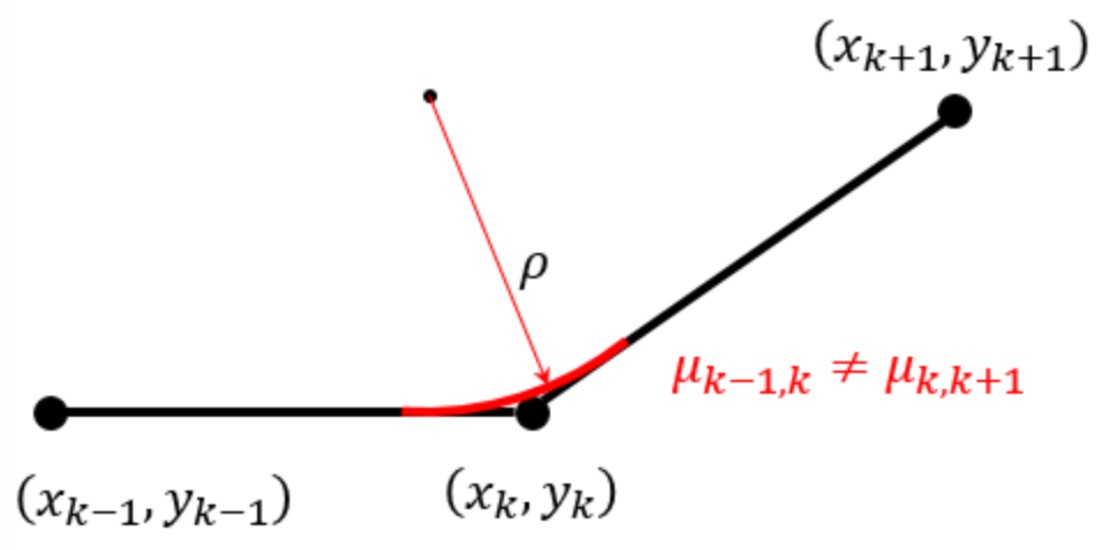}}
\caption{Desired vehicle paths given $\mu_{k-1,k}$ and $\mu_{k,k+1}$.}
\label{EgoCarPAth}
\end{figure}

\textbf{Remark}: We define a nominal speed $v_0$ for traversal along the desired path. If $\mu_{k-1,k}\neq \mu_{k,k+1}$, then $v_0$ should satisfy the following inequality:
\begin{equation}
    \dfrac{{v_0}^2}{\rho}\leq \dot{\psi}_{\mathrm{max}},
\end{equation}
where $\dot{\psi}_{\mathrm{max}}$ is the maximum yaw rate.

Given $\mu_{k-1,k}$ and $\mu_{k,k+1}$, one of the following two conditions holds:
\begin{itemize}
    \item{If $\mu_{k-1,k}= \mu_{k,k+1}$, then the projection of the desired path onto the $x-y$ plane is a single line segment connecting $(x_{k-1},y_{k-1})$ and $(x_{k+1},y_{k+1})$ (see Fig. \ref{EgoCarPAth}(a)).}
    \item{If $\mu_{k-1,k}\neq \mu_{k,k+1}$, then the projection of the desired path onto the $x-y$ plane consists of two separate crossing line segments connected by a circular path with radius $\rho$.  (See Fig. \ref{EgoCarPAth}(b).).}
\end{itemize}

\subsubsection{Trajectory Planning State Machine (TPSM)} 
A trajectory planning state machine (TPSM), shown in Fig. \ref{TPSM}, describes how the desired trajectory can be planned given vehicle (i) actual speed $v_T$, (ii) nominal speed $v_0$, (iii) turning radius $\rho$, (iv) maximum yaw rate $\dot{\psi}_{\mathrm{max}}$, and (v) consecutive path segment parameters $\mu_{{k-1},k}$ and $\mu_{{k},{k+1}}$. 
TPSM inputs are defined by the set
\[
\Sigma=\{v_T, \mu_{{k-1},{k}}, \mu_{{k},{k+1}}\}.
\]
TPSM states are defined by 
\begin{equation}
\mathcal{P}=\{p_0,p_1,p_2,p_3,p_4\},
\end{equation}
where atomic propositions $p_0$, $p_1$ and $p_2$ are assigned as follows:
\[
\begin{split}
    p_0:&{F}_N=\hat{\mathbf{k}}_B\cdot\left(mg\mathbf{k}_G+m\ddot{\mathbf{r}}\right)>0\\
    p_1:&~\mu_{{k},{k+1}}=\mu_{{k-1},{k}}\\
    p_2:&~v_T< v_0\\
     p_3:&~v_T= v_0\\
    p_4:&~\dfrac{v_0}{\rho}\leq \dot{\psi}_{\mathrm{max}}\\
\end{split}
.
\]
Note that $p_0$ is the initial TPSM state. TPSM terminal states are defined by the set
\[
\mathcal{F}=\{\mathrm{ACC},\mathrm{DEC},\mathrm{CV}\}
\]
where $\mathrm{ACC}$ and $\mathrm{DEC}$ command the car to accelerate and decelerate, respectively, and $\mathrm{CV}$ commands the car to move at constant speed.

Transitions over TPSM states are shown by solid and dashed arrows. If $p_k$ ($k=0,1,2,3,4$) is satisfied, transition to the next state is shown by a solid arrow; otherwise, state transition is shown by a dashed vector. 


\subsection{Motion Control}
\label{Feedback Control}
Suppose
\begin{equation}
\begin{split}
    \mathbf{r}_d=&x_{d}\hat{\mathbf{i}}_G+y_{d}\hat{\mathbf{j}}_G+f(x_d,y_d)\hat{\mathbf{k}}_T\\
\end{split}
\end{equation}
defines the desired trajectory of the car over the surface $\phi_p=z-f(x,y)=0$. 
Let $x$ and $y$ components of the car acceleration be chosen as follows:
\begin{equation}
\label{ddxddy}
\begin{split}
    \begin{bmatrix}
    \ddot{x}\\
    \ddot{y}
    \end{bmatrix}
    =
    \begin{bmatrix}
    \ddot{x}_{d}\\
    \ddot{y}_{d}
    \end{bmatrix}
    +k_1
    \left(\begin{bmatrix}
    \dot{x}_{d}\\
    \dot{y}_{d}
    \end{bmatrix}
    -
    \begin{bmatrix}
    \dot{x}\\
    \dot{y}
    \end{bmatrix}\right)
    +k_2
    \left(\begin{bmatrix}
    {x}_{d}\\
    {y}_{d}
    \end{bmatrix}
    -
    \begin{bmatrix}
    {x}\\
    {y}
    \end{bmatrix}\right)
\end{split}
.
\end{equation}
The error signal $
\mathbf{E}=\left(
\begin{bmatrix}
{x}\\
{y}
\end{bmatrix}
-
\begin{bmatrix}
{x}_{d}\\
{y}_{d}
\end{bmatrix}\right)
$
is then updated by the following second order dynamics:
\begin{equation}
\label{ERRRDYN}
    \ddot{\mathbf{E}}+k_1\dot{\mathbf{E}}+k_2\mathbf{E}=\mathbf{0}.
\end{equation}
The error dynamics is asymptotically stable and $\mathbf{E}$ asymptotically converges to $\mathbf{0}$ if $k_1>0$ and $k_2>0$.
Given $\ddot{x}$ and $\ddot{y}$ assigned by Eq. \eqref{ddxddy}, $\ddot{z}$ is specified by  
\begin{equation}
\label{GeomConst}
    \ddot{z}=\left(\dfrac{\partial f}{\partial x}\ddot{x}+\dfrac{\partial^2 f}{\partial x^2}\dot{x}^2+\dfrac{\partial^2 f}{\partial y^2}\dot{y}^2+\dfrac{\partial f}{\partial y}\ddot{y}+2\dfrac{\partial^2 f}{\partial x\partial y}\dot{x}\dot{y}\right).
\end{equation}
By knowing $\ddot{\mathbf{r}}$, the car control inputs $a_T=\dot{v}_T$ and $\gamma=\dot{\delta}$ are assigned by Eq. \eqref{atgamma}.

\section{Case Study Results}
\label{Simulation Results}
This section describes processing and infusion of map and weather data into our off-road multi-layer planner. In Section \ref{Global Route Planning}, data training and global route planning using dynamic programming are described. A local path planning example is provided in Section \ref{Local Path Planning}, and trajectory tracking results are presented in Section \ref{Trajectory Tracking}.

\subsection{Global Route Planning}


\subsubsection{Data Training}

The elevation data used for generating the grid-based map is downloaded from the United States Geographic Survey (USGS) TNM download \cite{USGSTNM}, Elevation Source Data (3DEP). The USGS elevation data is in ".las" format and needed to be transformed into a $1000\times 1000$ grid map. An online tool is used to transform the data into ".csv" format (see Ref. \cite{LidarTool}). After  data processing, we can obtain a map with raw elevation data for input to planning. 

Two different locations 
are chosen for this study. One is a mountainous area near the Ochoco National Forest, Oregon. The center location coordinate is $(44.2062527, -119.5812443)$ \cite{Mountain}. The second locale is in Indiana, near Lake Michigan, which is a relatively flat terrain area. The center location coordinate for the Indiana region is $(41.1003777, -86.4307332)$ \cite{Land}. 

\begin{figure*}[!ht]
 \centering
 \subfigure[]{\includegraphics[width=0.26\linewidth]{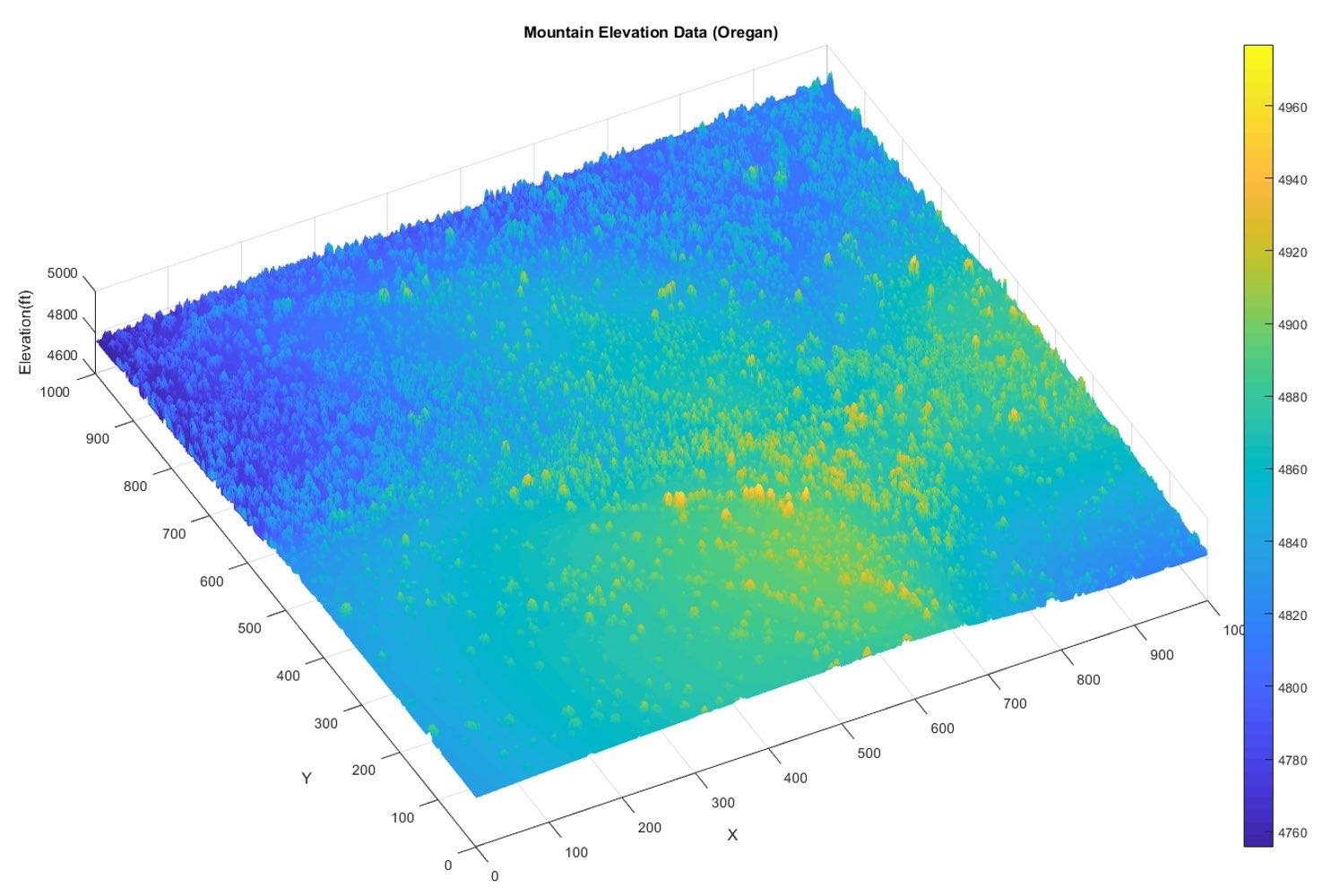}}
 \subfigure[]{\includegraphics[width=0.26\linewidth]{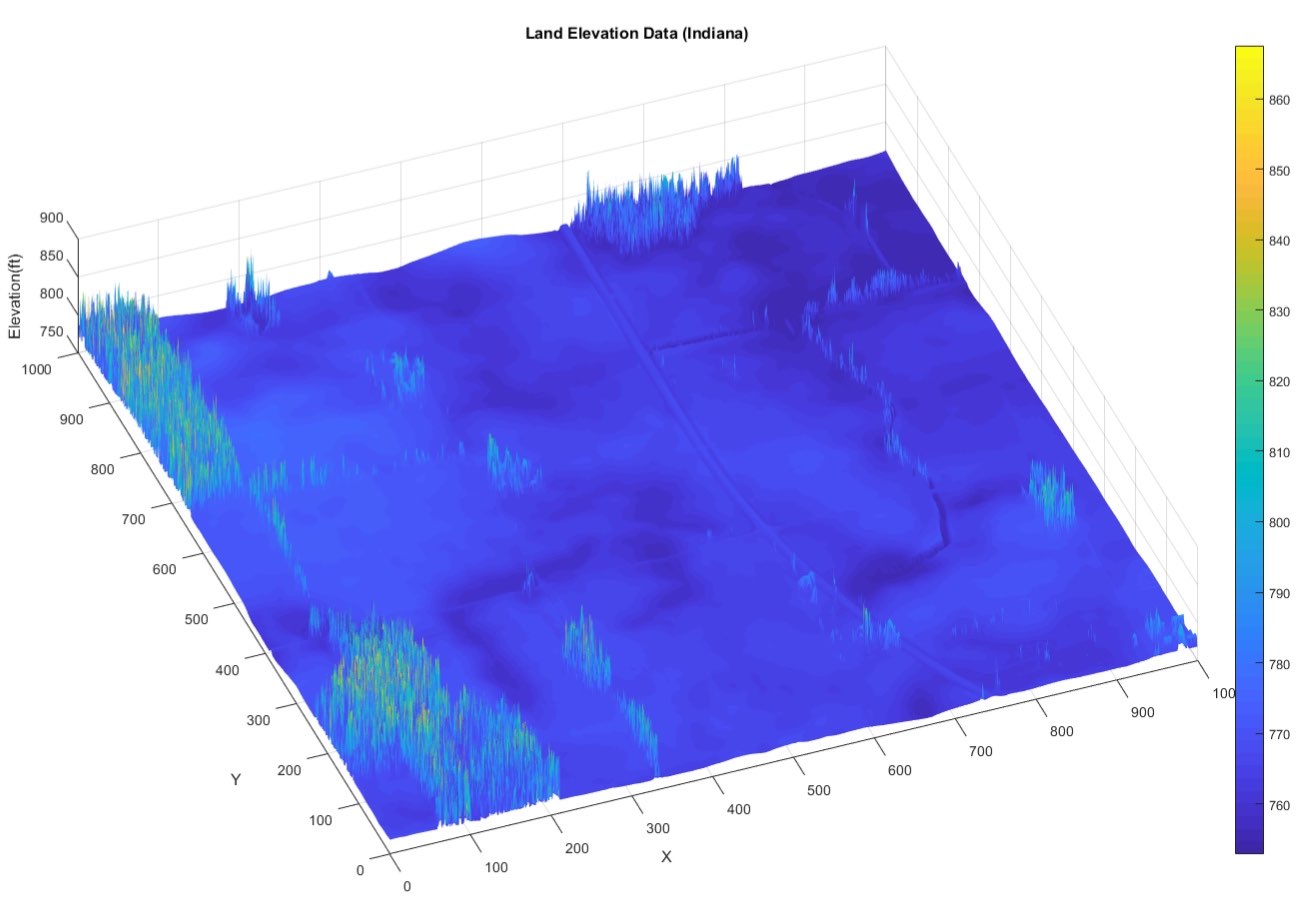}}
 \subfigure[]{\includegraphics[width=0.23\linewidth]{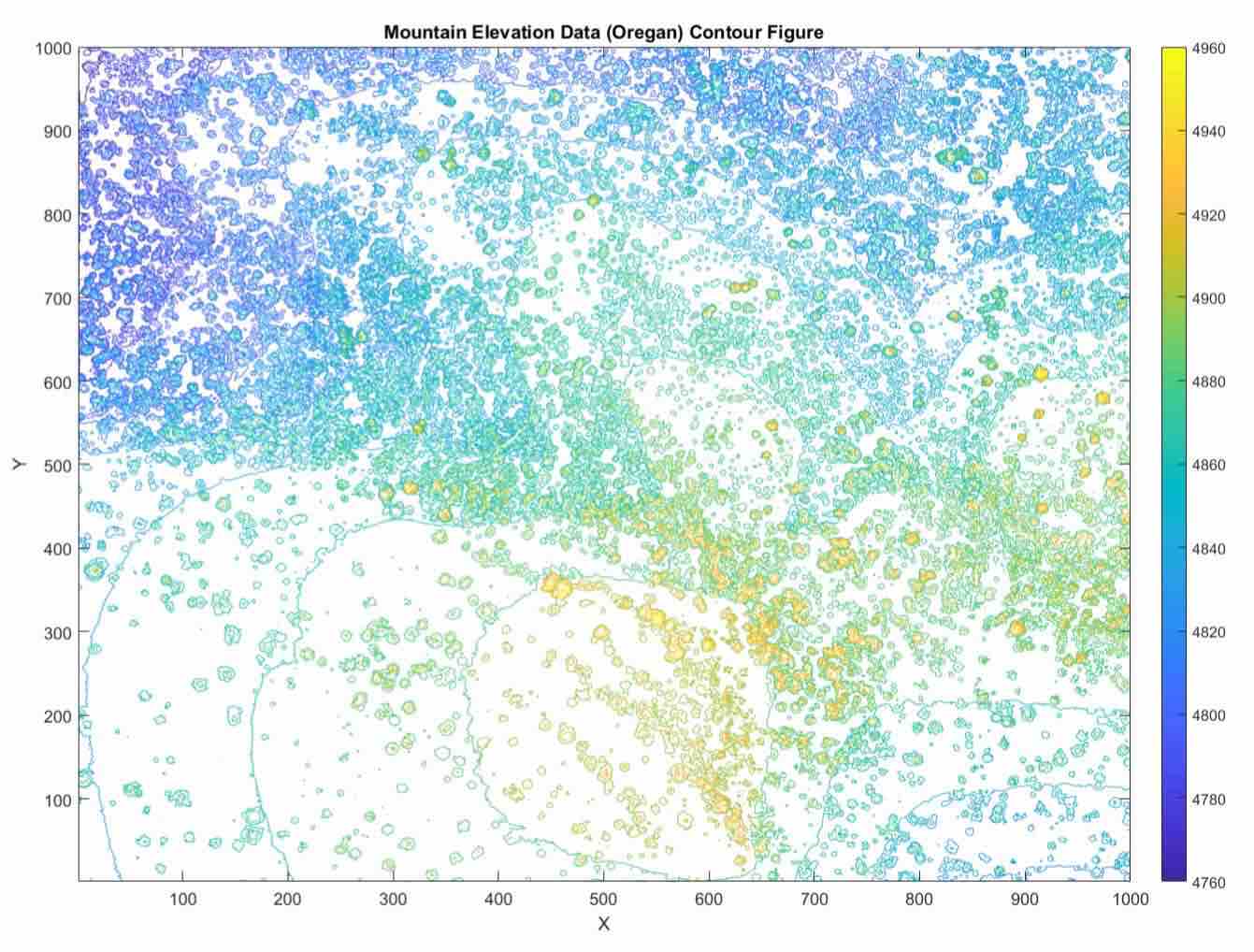}}
 \subfigure[]{\includegraphics[width=0.21\linewidth]{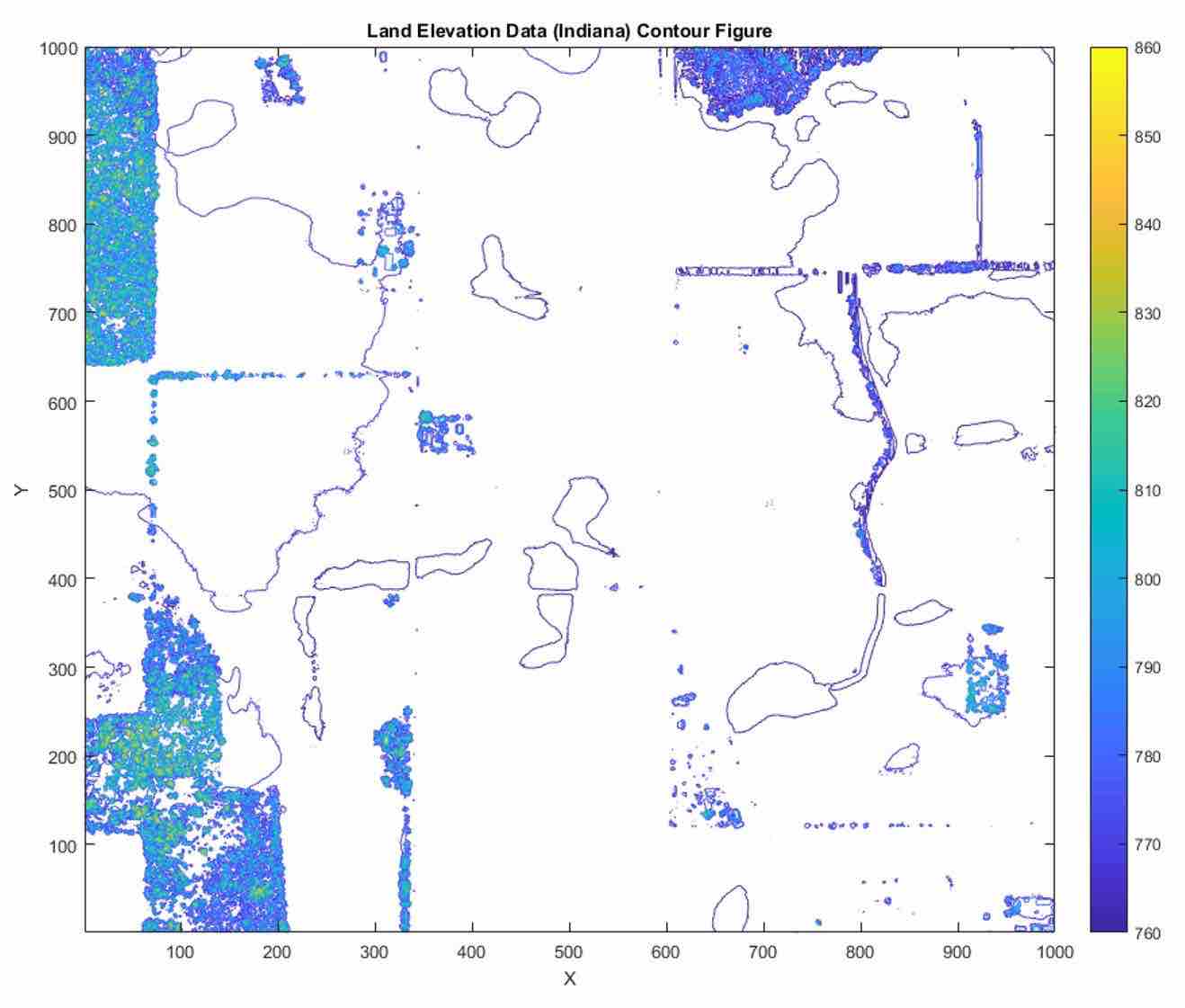}}
\caption{Elevation data for mountain (Oregon) and midwest (Indiana) case study terrains, with 3-D plots and contour plots: (a) Mountain Elevation Map (Oregon, coordinate (44.2062527, -119.5812443)). (b) Land Elevation Map (Indiana, coordinate (41.1003777, -86.4307332)). (c) Mountain Elevation Contour Map (Oregon, coordinate (44.2062527, -119.5812443)). (d) Land Elevation Contour Map (Indiana, coordinate (41.1003777, -86.4307332)).}
\label{SimulationMaps}
\end{figure*}

The $3-D$ surface maps and contour plots of both areas are shown in Fig. \ref{SimulationMaps}. The first Mountain area (Oregon Forest) has an average altitude of $4800$ feet and is covered by trees. The left lower area of the map has higher elevation and is covered with fewer trees. 

The second land area (Indiana) is flat with only several trees and roads as notable features. This area offers easier traversability in all weather conditions than the mountainous region.

The weather data are downloaded from a National Center for Environmental Information (NOAA) website.  Thee database contains temperature, weather type, and wind speed information. The weather is almost the same across each regions being traversed but it varies over time. If the weather is severe, such as snowy and rainy, the weather is called harsh or \textit{wet}. The constraint (threshold) on driving slope 
is decreased to $M_{w,max}$ under a wet weather condition. If the weather is dry, the slope constraint is set to $M_{d,max}$. 

By using the two typical land type elevation maps, global route planning simulation results are generated using DP. A $1000\times 1000$ grid map is obtained by spatial discretization of the study areas. A DP state $s\in \mathcal{S}$ represents a node in the grid map. As mentioned in Section \ref{Global Route Planning}, $9$ dicrete actions assign motion direction at a node $s\in \mathcal{S}$. Given $s\in \mathcal{S}$, the next waypoint $s_a\in \mathcal{S}$ given $a\in\mathcal{A}$ is considered unreachable if elevation change along the connecting path exceeds applicable upper-bound limit $M_{d,\mathrm{max}}$ or $M_{w,\mathrm{max}}$.
GRP case study results for Oregon and Indiana Maps are shown in Fig. \ref{MapPathPlan}. Three different destinations are defined in different GRP executions given the same initial location for each. Optimal paths connecting initial and final locations are obtained under nominal and harsh weather conditions as shown by  blue, red and green in Fig. \ref{MapPathPlan}. 

\subsubsection{Results For Nominal Weather Condition}
For nominal weather condition, we choose $M_{d,\mathrm{max}}=\tan (6.90^o)$ as the upper-limit (threshold) slope. Figs. \ref{MapPathPlan} (b) and (e) show corresponding optimal paths. Blue, red and green paths are reachable in both figures.
The heavily-forested Oregon area shown in Fig. \ref{MapPathPlan} impacts traversals. Except for the blue path starting from the edge of the forest, initial traversals are flat in the remaining paths. 

\subsubsection{Results For Harsh (Wet) Weather Conditions}
\begin{figure*}[!ht]
 \centering
 \subfigure[]{\includegraphics[width=0.27\linewidth]{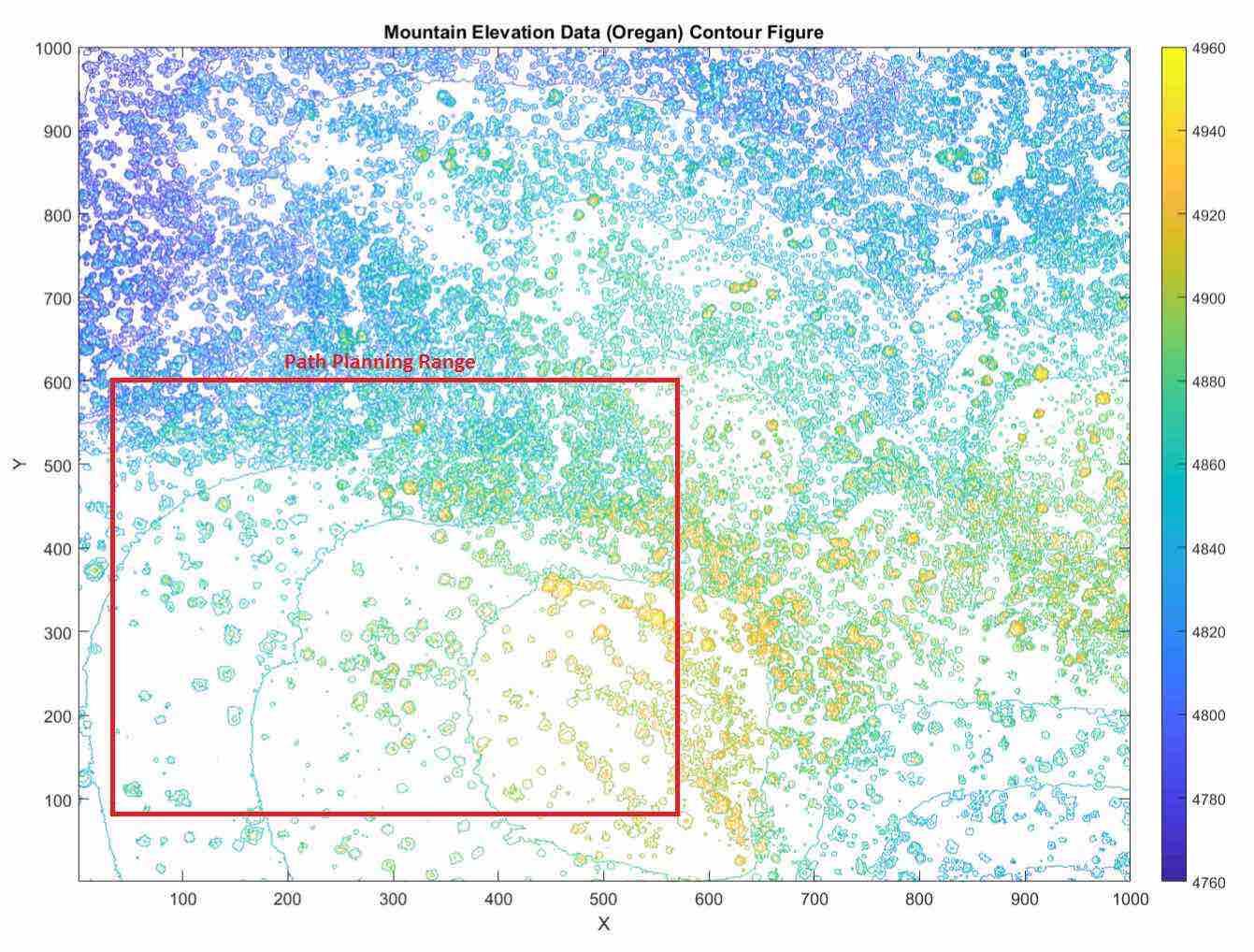}}
 \subfigure[]{\includegraphics[width=0.32\linewidth]{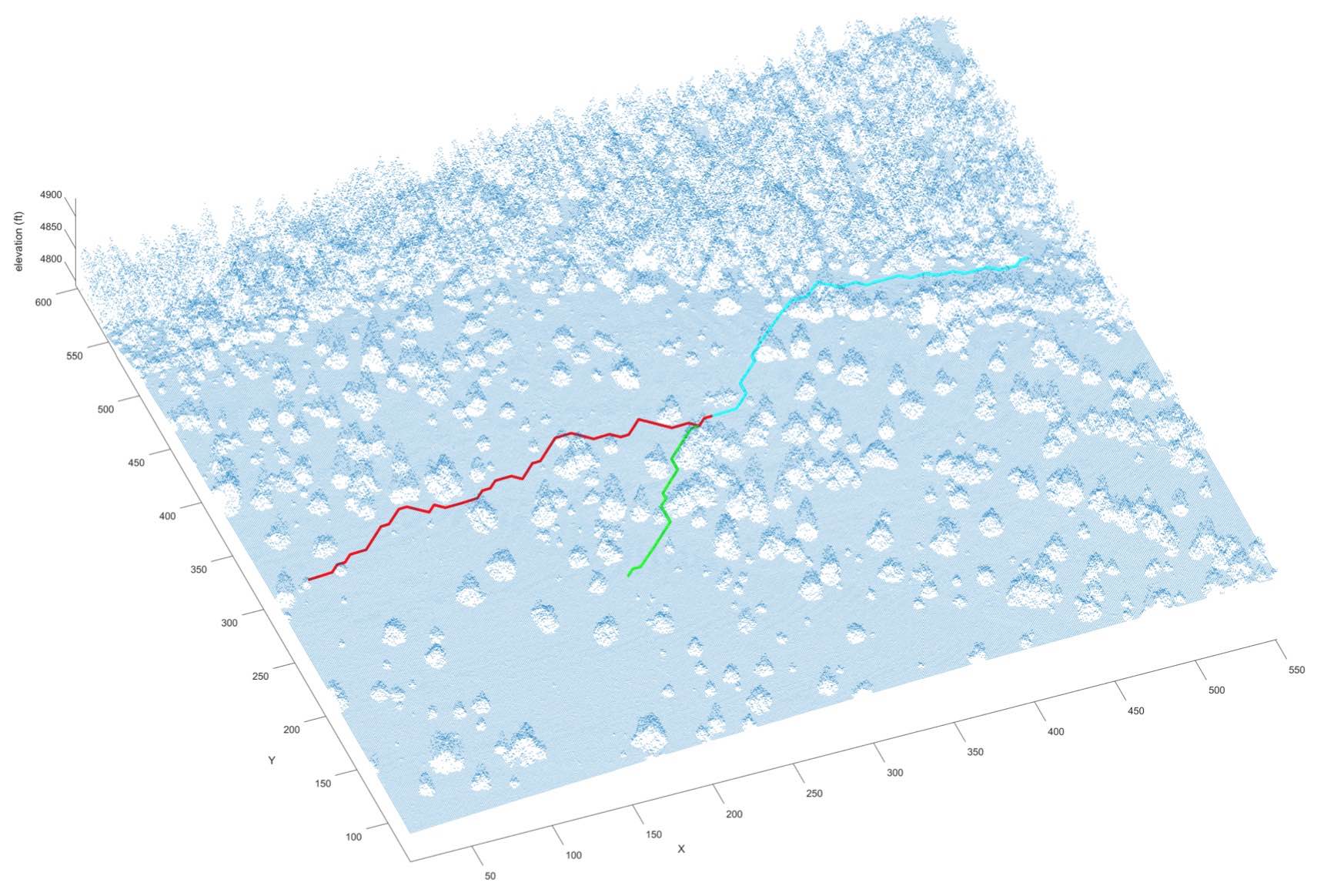}}
  \subfigure[]{\includegraphics[width=0.32\linewidth]{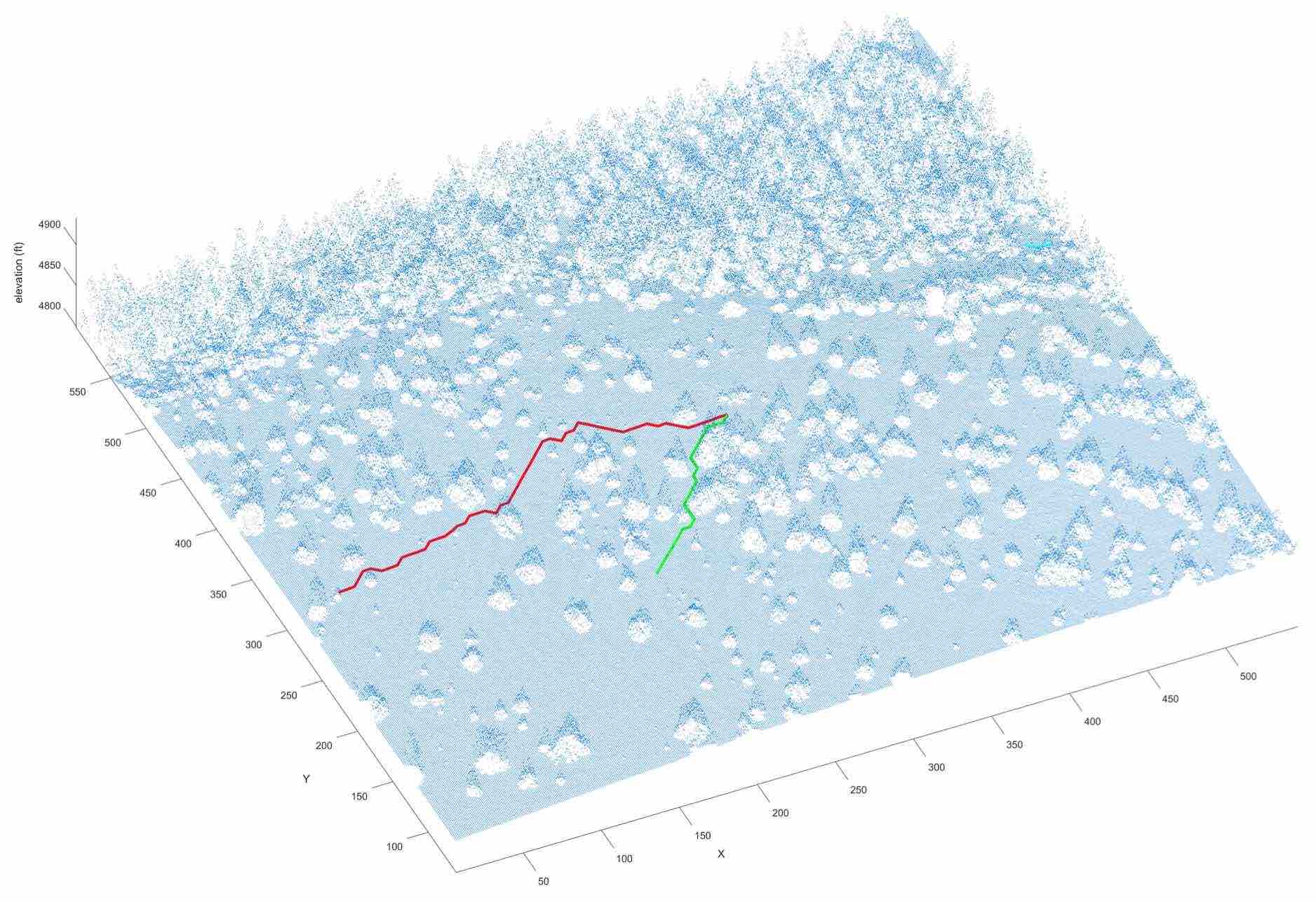}}
  \subfigure[]{\includegraphics[width=0.26\linewidth]{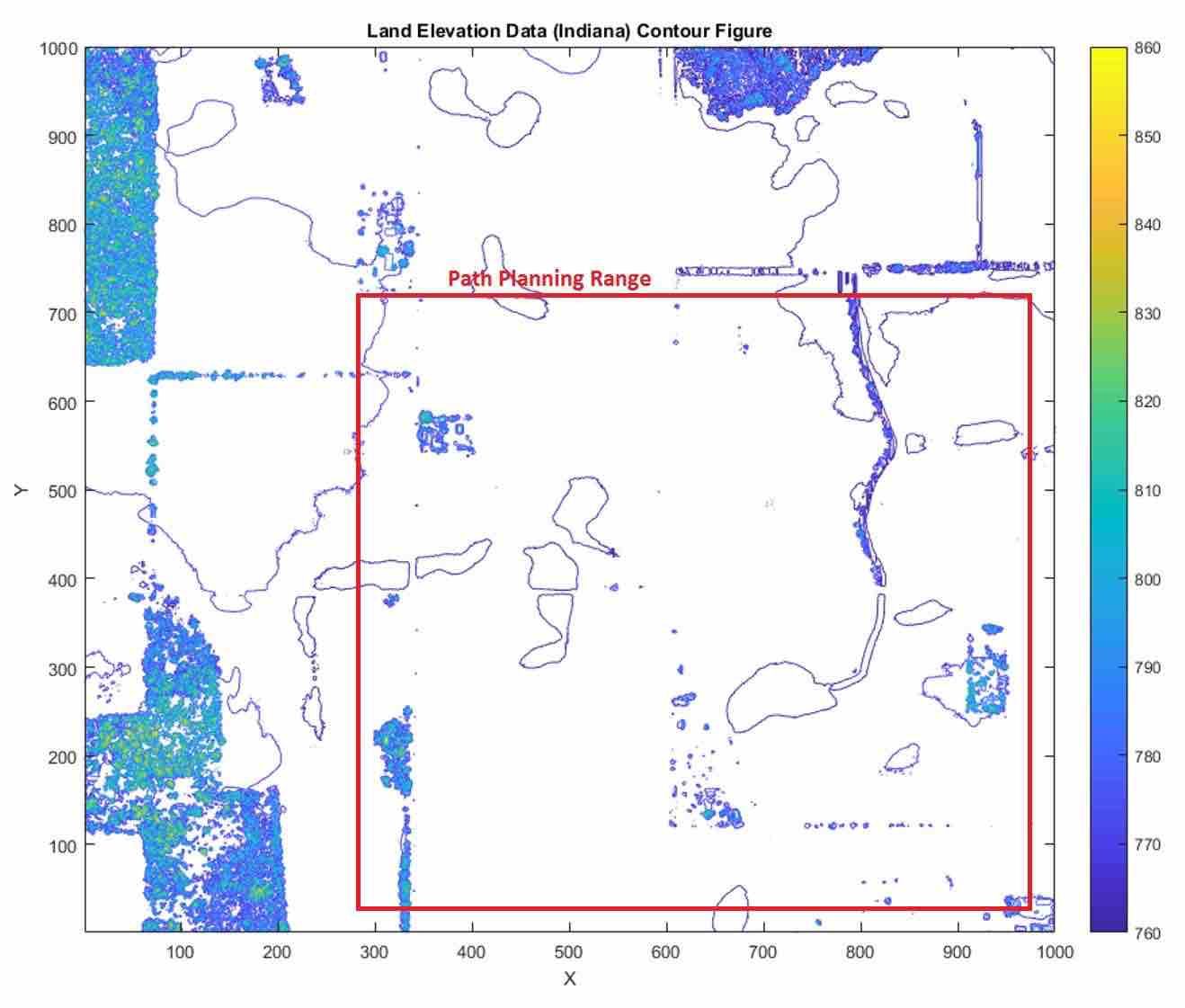}}
 \subfigure[]{\includegraphics[width=0.33\linewidth]{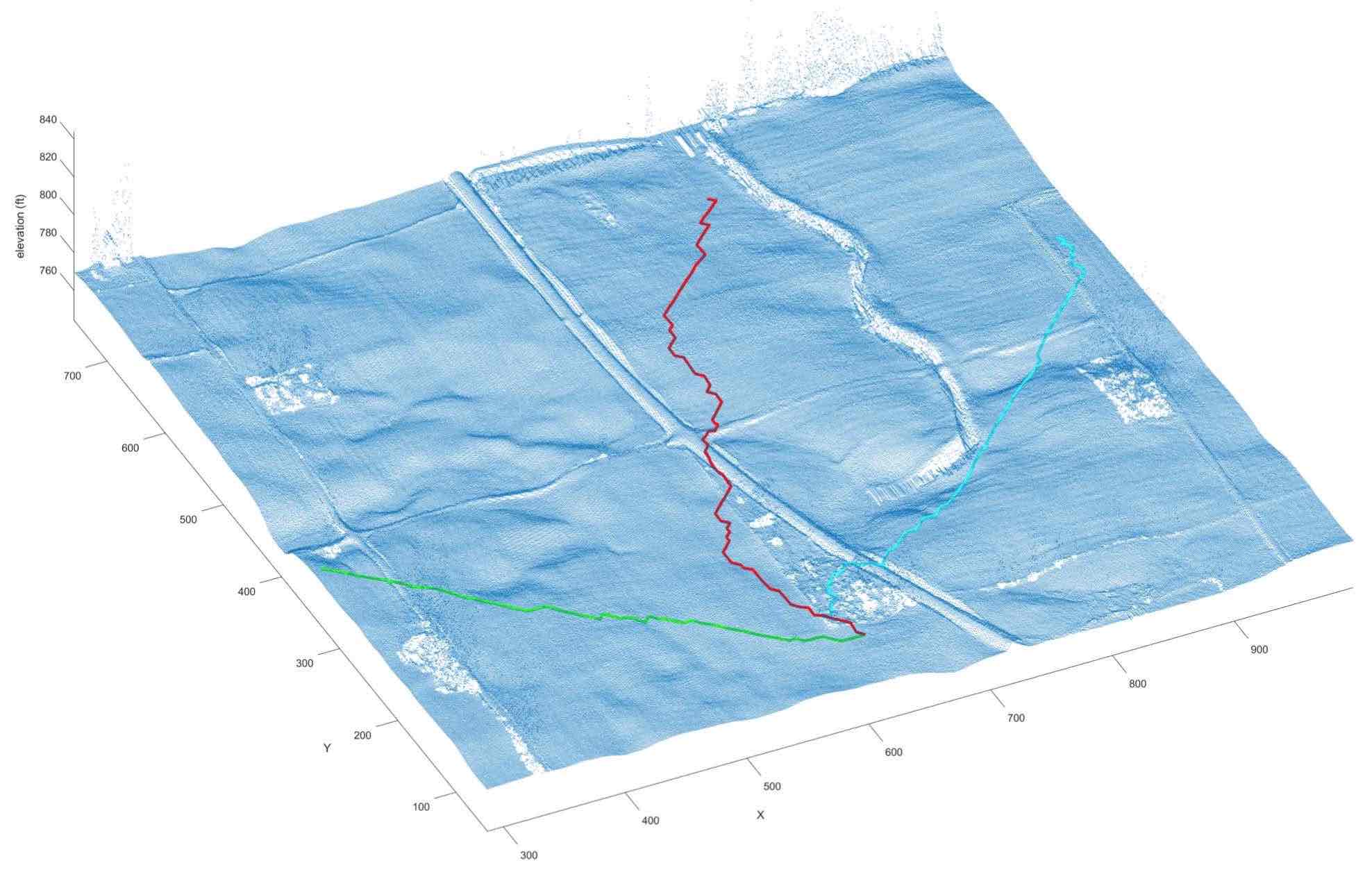}}
  \subfigure[]{\includegraphics[width=0.33\linewidth]{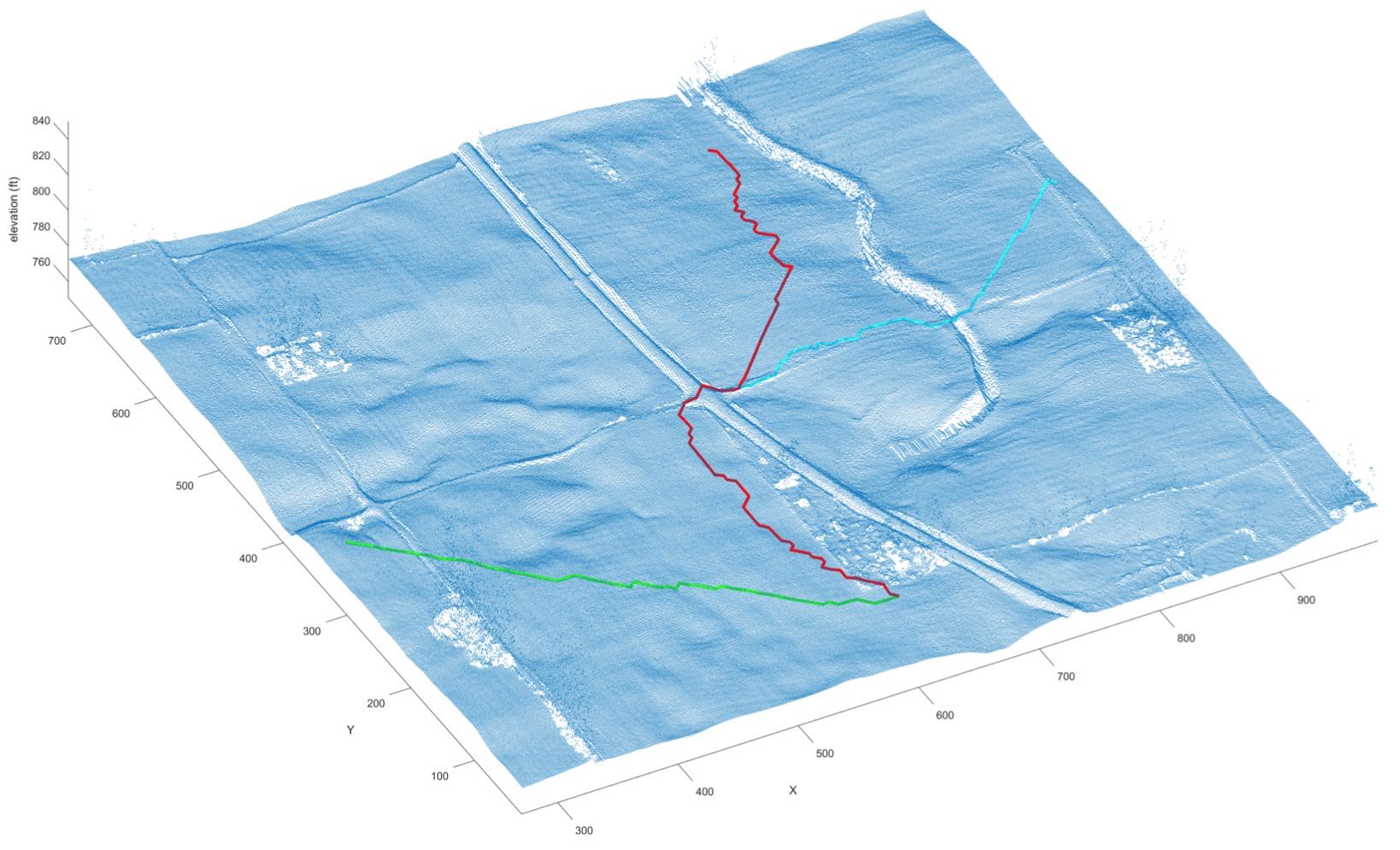}}
\caption{Three path plans are generated given the same start point with different constraint sets for the Oregon map and land map.  (a) The chosen traversal area for the Oregon map. (b) Nominal weather conditions for the Oregon map; the planned path has slope constraint $6.90^o$, representing $12.10\%$ slope. All paths are reachable. (c) Harsh (wet) weather for the Oregon map; the planned path has slope constraint $2.77^o$, representing $4.84\%$ slope. The blue path is unreachable while the other two paths are reachable given this constraint. (d) The chosen traversal area for the Indiana Map. (e) Nominal weather conditions for the Indiana map; the planned path has slope constraint $6.90^o$, representing $12.10\%$ slope. All paths are reachable. (f) Harsh (wet) weather for the Indiana map; the planned path has slope constraint $2.77^o$, representing $4.84\%$ slope. All paths are reachable.}
\label{MapPathPlan}
\end{figure*}

For harsh or wet weather conditions, we choose $M_{d,\mathrm{max}}=\tan (2.77^o)$ as the terrain slope constraint.
We consider the same start and target destinations to compute driving paths under harsh weather condition as in the previous cases. Figs. \ref{MapPathPlan} (c) and (f) show optimal driving paths under harsh (wet) weather conditions.  Note that in Fig. \ref{MapPathPlan} (c), the blue path destination is unreachable because the endpoint region is not connected to the center (start state) region due to terrain slope constraints.
The red and green paths are reachable but differ nontrivially compared to paths obtained for nominal weather conditions. As shown in Fig. \ref{MapPathPlan} (f),  GRP chooses a safer but longer path to avoid a low elevation region in the depicted bottom right region given bad (wet) weather. 
Path planning results under wet and nominal weather conditions are quantitatively compared in Table  \ref{table_example}.

\begin{table}[!ht]
\caption{Distances $D_{\mathrm{max}}(m)$, maximum and average slopes $\bar{s}( ^o )$ and $s(^o)$ of planned paths under nominal and harsh (wet) weather conditions.}
\label{table_example}
\begin{tabular}{| c | c | c | c | c | c | c | c |}
\hline
\multicolumn{2}{|c|}{}&\multicolumn{3}{|c|}{Appropriate}    & \multicolumn{3}{|c|}{Harsh} \\
  \hline
  State &Path&$D_{\mathrm{max}}$&$\bar{s}$&${s}_{\mathrm{max}}$&$D_{\mathrm{max}}$&$\bar{s}$&${s}_{\mathrm{max}}$\\
  \hline
\multirow{3}{*}{Oregon}& Red&$325.05$&$1.96$&$2.87$&$340.9$&$1.75$&$2.70$\\
& Green&$158.05$&$1.14$&$3.29$&$158.15$&$1.16$&$2.53$\\
  & Blue&$282.65$&$2.11$&$6.64$&$N/A$&$N/A$&$N/A$\\
  \hline
\multirow{3}{*}{Indiana}&Red&$680.1$&$0.92$&$5.54$&$728.4$&$0.75$&$2.67$\\
 & Green&$439.75$&$0.57$&$2.30$&$439.75$&$0.57$&$2.30$\\
& Blue&$637.4$&$1.11$&$6.33$&$780.5$&$0.73$&$2.63$\\
  \hline
 \end{tabular}
\end{table}

\begin{figure}[!ht]
\center
\includegraphics[width=3 in]{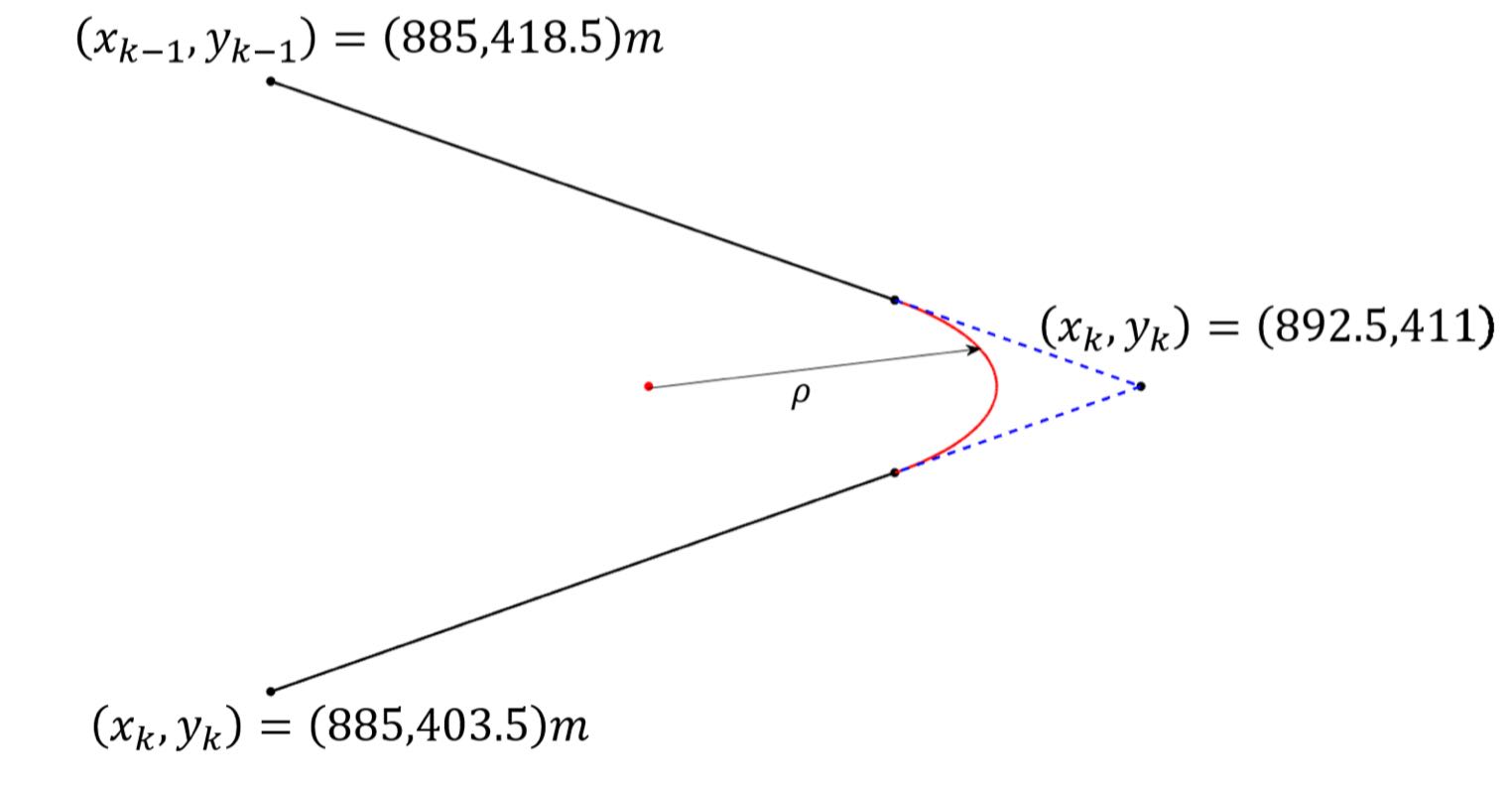}
\caption{Smooth turns computed during local path planning (LPP) given a turn-back defined by three consecutive waypoints.}
\label{LPPPic}
\end{figure}

\subsection{Local Path Planning}
\label{Local Path Planning}
Given three consecutive desired waypoints $(x_k,y_k)=(885,418.5)$ meters ($m$), $(x_k,y_k)=(892.5,411)$ m, and $(x_k,y_k)=(885,403.5)m$ , $\mu_{k-1,k}\neq \mu_{k,k+1}$. The desired path therefore consists of two straight path segments connected by a circular-arc turn (see Fig. \ref{LPPPic}). Note that the radius of the circular path is $\rho=4m$ in our case study. Selecting  $\dot{\psi}_{\mathrm{max}}=1~\mathrm{rad}/s$ as the upper-bound for yaw rate, atomic proposition $p_4$ is satisfied if $v_T=v_0=2~m/s$.
Because the desired speed is constant, the desired trajectory $\mathbf{r}_d(t)=x_d(t)\hat{\mathbf{i}}+y_d(t)\hat{\mathbf{j}}$ is given by
\begin{equation}
\label{Des1}
\begin{split}
    \mathbf{r}_d=&v_T\mathbf{\hat{n}}_d=2\mathbf{\hat{n}}_d\\
\end{split}
\end{equation}
where the tangent vector
\begin{equation}
\label{Des2}
    \mathbf{\hat{n}}_d=\dfrac{d\mathbf{r}_d}{ds}
\end{equation}
is as follows:
\begin{equation}
\label{Des3}
    \mathbf{\hat{n}}_d=
    \begin{cases}
   {\frac{\sqrt{2}}{2}}\hat{\mathbf{i}}+{-\frac{\sqrt{2}}{2}}\hat{\mathbf{j}}&s\leq  4.6066\\
    2\cos\left(-s+{\frac{\pi}{4}}\right)& 4.6066<s\leq 14.0314\\
    {-\frac{\sqrt{2}}{2}}\hat{\mathbf{i}}+{-\frac{\sqrt{2}}{2}}\hat{\mathbf{j}}&14.0314<s\leq 18.6380\\
    \end{cases}
    .
\end{equation}
Note that $0\leq s\leq 18.6380$ is the desired path arc length, where $v_T=\dot{s}=2{~m/s}$ ($\forall s$).

\subsection{Trajectory Tracking}
\label{Trajectory Tracking}
\begin{figure}
\center
\includegraphics[width=2.8 in]{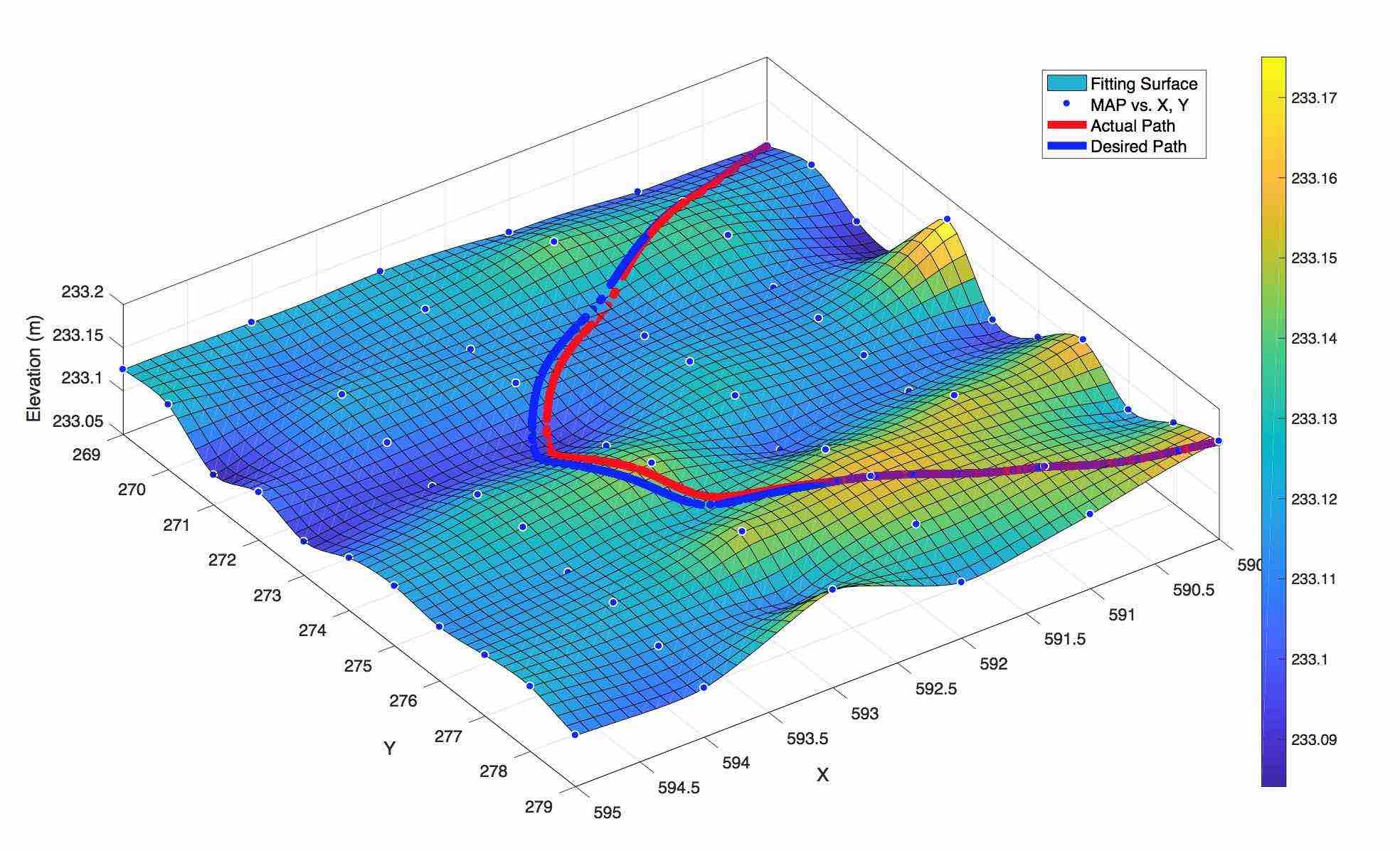}
\caption{Actual and desired trajectory of the car over the motion surface}
\label{Coefficients}
\end{figure}

\begin{figure}[!ht]
\center
\includegraphics[width=3 in]{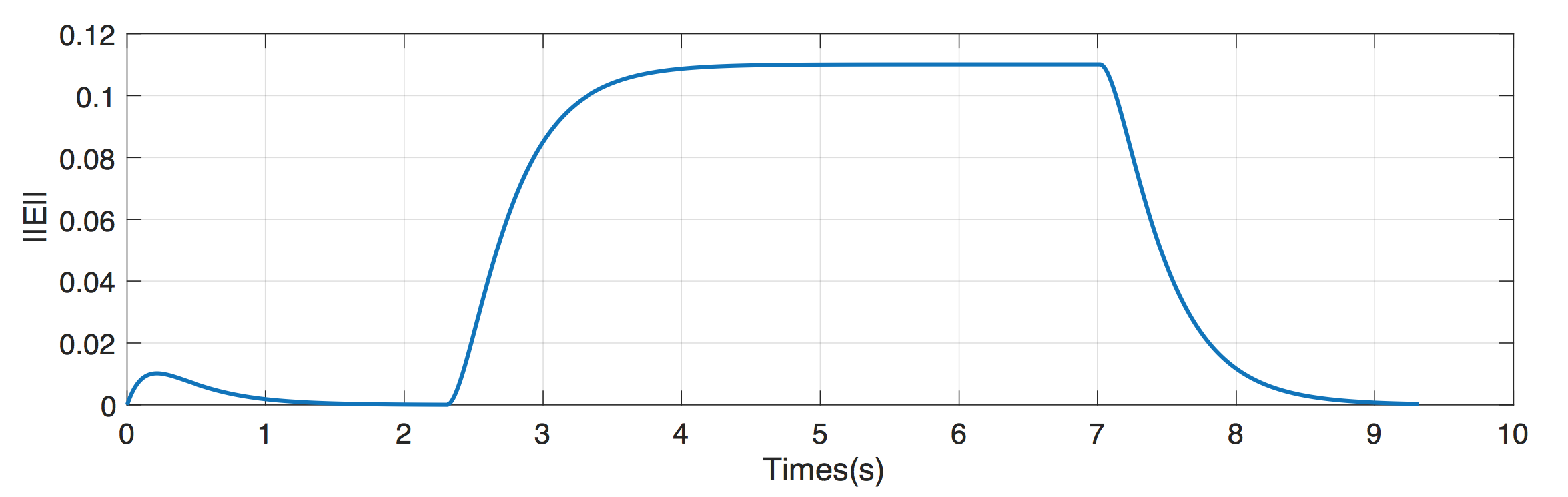}
\caption{Deviation between actual and desired position of the car as a function of time}
\label{EERROORR}
\end{figure}

By applying the proposed feedback controller design from Section \ref{Feedback Control}, the desired LPP vehicle trajectory assigned by Eqs. \eqref{Des1} and \eqref{Des2} can be asymptotically tracked. The FC is assigned controller gains $k_1=10$ and $k_2=20$ for this example.
 Fig. \ref{Coefficients} shows the $x$ and $y$ components of vehicle desired and actual positions. Error $\|\mathbf{E}\|$, the deviation between actual and desired position, is shown versus time in Fig. \ref{EERROORR}. Notice that error never exceeds $0.1101 $. This deviation error occurs due to (i) surface non-linearities and (ii) sudden acceleration changes along the desired path. While acceleration along the linear segments of the path is zero, acceleration rapidly changes when the car enters or exits a circular arc (turning) path.

\section{Conclusion}
\label{Conclusion}
This paper presents a novel data-driven approach for off-road motion planning and control. A dynamic programming module defines optimal waypoints using available GIS terrain and recent weather data. A path planning layer assigns a feasible desired trajectory connecting planned waypoints. A feedback linearization controller successfully tracks the desired trajectory over a nonlinear surface.  In future work, we will relax assumptions related to sideslip and incorporate more sophisticated models of terrain interactions to improve decisions across all three decision layers.

\section{\hspace{0.2cm} Acknowledgement}
This work was supported in part under Office of Naval Research grant N000141410596. 

\bibliographystyle{IEEEtran}

\end{document}